\documentclass[conference]{IEEEtran}
\usepackage{cite}
\usepackage{amsmath,amssymb,amsfonts}
\usepackage{graphicx}
\usepackage{tabularx}
\usepackage{booktabs}
\usepackage[hidelinks]{hyperref}  
\usepackage{xurl}
\IEEEoverridecommandlockouts

\begin{document}

\title{Forms of LLM-Integrated Applications  from LLM-Chats to Autonomous AI Agent Systems}

\author{
  \IEEEauthorblockN{Irene Weber}
  \IEEEauthorblockA{\textit{Faculty of Mechanical Engineering},
    \textit{University of Applied Sciences Kempten},
    Kempten, Germany \\
    Email: irene.weber@hs-kempten.de}
  \thanks{Accepted at 4th International Conference on Foundation and Large Language Models (FLLM2026), 17--20 November 2026, Barcelona, Spain.}
}

\maketitle

\begin{abstract}
Large language models (LLMs) are increasingly embedded as components in
software systems, marketed under labels such as chatbot, copilot,
retrieval-augmented generation, workflow, coding agent and AI agent. Whether
these labels denote genuine architectural forms or serve as branding has not
been assessed systematically.

In the sources surveyed, labels do carry architectural content, most clearly in
vendor usage: copilot denotes a router-worker architecture operating a host
application under step-by-step user confirmation, while the more recent shift to
the label agent coincides with AI-planned multi-step execution of which the user
sees only the outcome. The coding agents of four major providers share one
architecture, a reason-and-act loop delegating to subagents.

This survey describes seven recurring forms---LLM chats, custom agents,
retrieval-augmented generation (RAG), AI-enhanced workflows, copilots, coding
agents, and, in part, agentic RAG---in a common vocabulary of agents and tools.
Each is characterized along four structural dimensions (agentic RAG   only partially): the architectural
pattern, the control of execution and the point of user intervention, the
number of agent calls per task, and tool use. An illustrative corpus of 22 systems from research
publications and vendor documentation grounds the descriptions and shows where
they reach their limit.
\end{abstract}


\section{Introduction}

Large language models (LLMs) have moved beyond their initial role as conversational assistants. They are used as components inside software systems, where their output is consumed not by a human reader but by other pieces of software. Embedded in a program, an LLM no longer serves only as a content generator: it selects actions, parametrizes external services, and influences what happens next.

A vocabulary for these systems has emerged along with them. Practitioners speak of chatbots, copilots, retrieval-augmented generation, workflows, AI agents and agentic AI, and vendors use the same terms in product names and branding. The terms are evidently useful, since they are used routinely and widely. What is missing is an account of what they denote: developer guides, engineering blogs and vendor material or product documentation describe what individual systems do, but rarely how they are built or how one form compares architecturally with another. LLM-integrated systems developed in research usually originate in an application domain, where this terminology is adopted rather than examined. Several of the terms appear to have been coined in the development and research laboratories of leading vendors and adopted by the academic literature afterward, rather than the reverse.

It remains unclear whether the naming is largely arbitrary or whether these terms denote distinct categories of artifact and, if so, in what respects they differ.
The result is a field that is hard to survey for newcomers and practitioners alike, and in which researchers struggle to position their work and compare it with related approaches.

This paper responds with a systematic description of the established
forms in a single, common vocabulary. Its primary contribution is one
of terminology and of conceptual groundwork, and only secondarily one
of analysis. In detail, the contributions are:
\begin{itemize}
	\item a set of concepts---AI agent, tool, monofunctional and
	multifunctional agent, and recurring architectural patterns---in
	which LLM-integrated applications can be described at the level of
	whole system architectures;
	\item uniform descriptions of seven forms established in practice but
	not yet defined in a common vocabulary, each characterized along four
	structural dimensions: architectural pattern, control of execution
	and the point at which the user intervenes, number of agent calls per
	task, and tool use;
	\item an overview of 22 systems from research publications and vendor
	documentation, which illustrates the descriptions, shows where they
	reach their limit, and shows that the label copilot denotes one
	architecture and one set of characteristics, while the shift to the
	label agent marks a shift in architecture.
\end{itemize}

Section~\ref{sec:related} reviews related work.
Section~\ref{sec:blocks} defines the concepts used throughout.
Section~\ref{sec:types} describes the seven forms.
Section~\ref{sec:systems} surveys selected systems.
Section~\ref{sec:discussion} discusses implications, limitations and
further work.

\section{Related Work}
\label{sec:related}

Conceptual work published by vendors presents catalogues of architectural
patterns intended as guidance for developers building LLM-based systems
\cite{anthropicBuildingEffectiveAgents2024,googleLLMAgents2025,openaiPracticalGuideBuilding2025}.
Surveys of multi-agent systems (MAS) address collaboration mechanisms
\cite{tranMultiAgentCollaborationMechanisms2025}, application domains
\cite{chenSurveyLLMbasedMultiAgent2025}, or the conceptual distinction
between AI agents and agentic AI \cite{sapkotaAIAgentsVs2026}. Masterman
et al.\ and Zhou et al.\ classify agent systems as single- versus
multi-agent and hierarchical versus decentralized architecture, but draw their examples from research prototypes
\cite{mastermanLandscapeEmergingAI2024,zhouSurveyArchitecturesAgentic2026}.
None of these works characterizes applications systematically.

In earlier work, the author proposed a taxonomy of LLM-integrated
applications that characterizes the individual model invocation, termed LLM
component, which corresponds to the notion of AI agent used in this paper.
That approach decomposes LLM-integrated applications into LLM components and
classifies each component along several dimensions
\cite{weberLargeLanguageModels2024a}, but stops short of describing whole
system architectures.

Closest to the present work is the taxonomy of H\"andler
\cite{handlerTaxonomyAutonomousLLMPowered2023}, which classifies autonomous
LLM-powered MAS by their levels of autonomy and alignment, assessed per
architectural aspect. Its alignment axis is a finer-grained treatment of what
appears here as user control; its autonomy axis is related to our number of
agent calls per task. It covers one class of systems, applied to seven
open-source projects, and includes a workflow automation tool expressly as a
contrasting case; the present paper describes a broader range of forms in
uniform terms. To the author's knowledge, no other work describes the range of
established forms in a single, common vocabulary at the level of application
systems.

\section{Agents and Tools}
\label{sec:blocks}

The descriptions in Section~\ref{sec:types} refer to systems which 
are assembled from smaller units. This section defines the terminology
adopted here.

\subsection{Invocation and Tools}

\begin{figure}[t]
  \centering
  \includegraphics[width=0.36\textwidth,alt={Diagram showing a single LLM invocation: a prompt flows into a box labeled LLM, which produces an output.}]{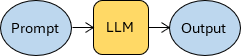}
  \caption{figure}{Invocation of an LLM.}
  \label{fig:llm-query}
\end{figure}

\begin{figure}[t]
  \centering
  \includegraphics[width=0.48\textwidth,alt={Diagram showing the control and data flow of tool use: a prompt flows into an LLM, which produces a tool call; the tool call is executed and its result is fed back into a second LLM step, which produces the final output.}]{  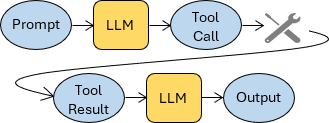}
  \caption{figure}{Control and data flow of tool use.}
  \label{fig:tool-use}
\end{figure}

An LLM is used by invoking it with a prompt, as shown in
Fig.~\ref{fig:llm-query}. A prompt comprises several types of information: an
instruction or system prompt, which defines the general behavior of the
prompted LLM, such as  ``act as a helpful assistant''; a state or context part,
which informs the LLM about the current circumstances, such as a dialog
history; and the task it has to solve, such as a user's question it has to
answer \cite{weberLargeLanguageModels2024a}.
An LLM cannot act; it can only produce text. Its scope is extended by giving
it access to \emph{tools}. 
The model uses a tool by emitting a call with parameters appropriate to the
task; the surrounding system executes it, holding the
function implementations, the service endpoints and credentials, or the
execution environment for generated code, and the result is optionally fed
into a further invocation from which the model formulates a response
(Fig.~\ref{fig:tool-use}).

For the model to use a tool correctly, the prompts must describe when the tool
applies and how it is called. Tool integration has begun to standardize: the
Model Context Protocol defines a uniform format in which services describe the
functions they offer and how a model may call them
\cite{anthropicModelContextProtocol2025}.

\subsection{AI Agent}

The terminology in the field of agentic AI has not yet converged. Some
authors call any prompt-and-response component an agent, for instance, in the context of custom agents in Subsection~\ref{sec:customagents}. Others require
tool use \cite{microsoftAgents2025}, planning capability
\cite{mitchellFullyAutonomousAI2025a}, or a degree of autonomous control over the
flow of execution \cite{roucher2024smolagents,googleLLMAgents2025}.
\emph{Agentic AI} is treated by some as an umbrella term that includes
single-agent systems \cite{anthropicBuildingEffectiveAgents2024} and by others as a
synonym for MAS \cite{sapkotaAIAgentsVs2026}. 

This paper adheres to a minimal definition: \textbf{an AI agent is a
prompt-and-response system component, with or without access to tools.}
The term \emph{LLM agent} would be more precise, but \emph{AI agent} is
adopted because it has become established and is consistent with the
equally established term \emph{agentic AI}. Under this definition, agents are the building blocks of agentic AI and of LLM-based software systems, i.e., systems that invoke LLMs.

\subsection{Monofunctional and Multifunctional Agents}

What constitutes a given AI agent is its prompt, above all its
instruction or system prompt. It defines the agent's behavior, while the
model that processes the prompt is in principle interchangeable. Agents can be designed with varying degrees of complexity.
A \emph{multifunctional} agent performs a range of
functions, selects the appropriate one itself and decides on its
execution; typically a whole palette of tools is at its disposal, which
it selects and parametrizes autonomously.
A \emph{monofunctional}
agent fulfils only a delimited function, for example parametrizing a
single tool.  
There is a continuum from strictly monofunctional to multifunctional agents, rather than a strict division. 

\subsection{Architecture Patterns}
\label{sec:amas}
LLM calls can be combined and integrated into systems in countless ways. Among these, certain basic architectures are---with varying names---described in the literature \cite{anthropicBuildingEffectiveAgents2024,mastermanLandscapeEmergingAI2024,zhouSurveyArchitecturesAgentic2026,openaiPracticalGuideBuilding2025}.
The following list names some of them.

\textbf{Monofunctional single agent} and \textbf{multifunctional single agent} (Fig.~\ref{fig:llm-query}), which may include  tool use, have been described in Section~\ref{sec:blocks}.
In the context of AI agents, the term \textbf{workflow} refers to applications in which the agents are embedded in a prescribed process. Workflows solve tasks over a sequence of process steps. Individual steps may be realized as agents, and the process supplies each agent invocation with data from the process context. The sequence is defined in advance and may contain conditional branches, but it is not composed dynamically by a model. A special case is the \textbf{prompt-chaining workflow}, consisting of a sequence of LLM calls, where the result of an LLM call is fed into the prompt of the next LLM call.
The \textbf{router-worker} architecture consists of a central router agent and a set of specialized worker agents. A worker is a subagent, typically monofunctional. An input to the system first reaches the router, which determines which of the system's predefined tasks the input matches---intent detection---and then invokes the worker agent responsible for the detected intent with parameters extracted from the system environment and the input. 

In the \textbf{orchestrator-worker} architecture a central manager agent plans and steers. It decomposes a task into subtasks and hands them to specialized workers by configuring their prompts; workers may run in parallel, and a further agent may consolidate the partial results.

The \textbf{planner-executor} architecture, also called manager-worker, adds iterative refinement. The planner produces a work plan, workers execute subtasks and additionally return feedback, and the planner revises the plan and delegates again until the desired result is reached. The \textbf{hierarchical planner-executor} variant inserts intermediate planners: a high-level planner performs coarse planning and passes subtasks to subordinate planners that refine them, with workers executing at the leaves. All architectures involving calls to workers or subagents have analogous hierarchical variants.

In the \textbf{reason-and-act (ReAct) loop}, a single multifunctional agent interleaves reasoning with tool calls in a loop: it reasons about what to do next, invokes a tool, evaluates the result in a further reasoning step, and repeats until it judges the task complete \cite{yaoReActSynergizingReasoning2023}. If the ReAct agent's tools include subagents, it constitutes a further MAS architecture.

These architectures vary in autonomy, that is, the degree of control left to an LLM within an agentic system. Some authors reserve the term autonomous MAS (AMAS) for architectures such as orchestrator-worker, reason-and-act, and planner-executor, where the number of LLM calls the MAS executes to fulfil a task is determined by an LLM rather than fixed, as in workflows or the router-worker architecture \cite{anthropicBuildingEffectiveAgents2024}.
Agent systems may go further still, writing automation scripts and defining prompts for subagents by themselves, thereby attaining far-reaching autonomy \cite{mitchellFullyAutonomousAI2025a}.
Architectures without central control are also possible, notably peer-to-peer architectures in which agents divide the task among themselves and coordinate through mechanisms such as shared blackboards, voting or auctions. 
In practice, centrally controlled architectures dominate in 2025: among published case studies of deployed MAS on one framework platform, the orchestrator-worker and router-worker architectures are the most frequent \cite{weber2025langchain}.

Agent systems may contain further agents that do not contribute to solving
the task but supervise it. Guardrail agents, for example, check the inputs
and outputs of other agents against safety rules, access rights or
compliance requirements and block or modify actions that violate them
\cite{xiangGuardAgentSafeguard2025}. Such agents are a cross-cutting
concern, comparable to logging or authorization in conventional software:
they can be added to any of the architectures listed above without altering
it. They are therefore not treated here as architectural elements, and their
invocations are not counted among the agent calls per task.

\section{Forms of LLM-Integrated Applications}
\label{sec:types}

This section describes established forms of LLM-integrated applications
by examples. 

\subsection{LLM Chats}\label{sec:chat}

LLM chats offer a dialogue interface to a model. In organizations they
are used for drafting texts of all kinds and for translation, and as a
convenient and powerful substitute for search engines, with the known
caveats that they occasionally produce fabricated information and that
their knowledge does not extend to the present.
Chat products typically work with multifunctional prompts, see for instance
\cite{schwartzChatGPTJustAccidentally2024}.  Users may request tool
use explicitly, or the model decides for itself. 

Behind a current chat product there may be
several models of different sizes together with a router that selects one
per user turn \cite{openaiIntroducingGPT52025}. 
The deep research
function offered by several chat products may be viewed as a tool within
the chat system, see Subsection~\ref{sec:rag}.  

\subsection{Custom Agents}\label{sec:customagents}

Effective prompts are often extensive and hard to develop, since small
changes in wording can substantially change their effect. It therefore
pays to store successful prompts for recurring tasks. Several AI chat
platforms allow such prompts to be managed as \emph{custom agents} in
their interface, and vendor-independent platforms offer the same across
models \cite{quorainc.PoeFastHelpful2025}. The names for custom agents differ by
platform. Often, tools can be attached, and some services operate
marketplaces in which custom agents are published and reused.

An analysis of the custom agent landscape found that external application
integration remains comparatively rare, with most agents relying on
capabilities the host service provides \cite{zhaoGPTsWindowShopping2024}, and guidance
for the format recommends scoping each agent to a delimited task
\cite{openaiacademyBuildingCustomGPTs2025}. Custom agents also occur inside larger
systems: workflows use them as process steps \cite{n8nBuildCustomAI2026}.

\subsection{Retrieval-Augmented Generation}\label{sec:rag}
Retrieval-augmented generation (RAG) \cite{lewis2020retrieval} preserves
the usability of the chat interface while opening knowledge sources that
did not enter model training, such as an organization's internal documents or specialized domain expertise. 
RAG retrieves passages from the knowledge sources to generate answers, which reduces fabrication
(hallucination), and makes the supporting documents or entries
explicit, so that an answer can be checked against them
\cite{kenthapadiGroundingEvaluationLarge2024,rashkinMeasuringAttributionNatural2023}.
RAG is among the most successful applications of LLMs in professional
settings. The following sections discuss two variants: RAG in the narrow
sense, and agentic RAG.

\subsubsection{Narrow Sense RAG}
RAG in the narrow sense goes back to \cite{lewis2020retrieval} and is
suited to large, heterogeneous and largely unstructured document
collections, such as manuals, project reports or case records built up
over years. In the form common today, the collection is prepared in
advance: text is extracted from the source formats, split into chunks,
converted into vectors by an embedding model and stored in a vector
database. 
At query time, the same embedding model converts the user's
question into a vector, and the system retrieves the chunks whose
vectors are most similar. These are inserted into the prompt, from which
an LLM then generates the answer.

The LLM plays no part in retrieval.
Thus, in the terminology adopted here, this is technically not an agent with a
tool as in Fig.~\ref{fig:tool-use}, but the agent is reduced to
formulating an answer from material handed to it as in
Fig.~\ref{fig:llm-query}.

\subsubsection{Agentic RAG}
Agentic RAG denotes approaches for searching large and frequently changing content such as company websites, intranets, wikis, document collections or databases. Retrieval is performed by appropriate mechanisms invoked as tools, such as web search, enterprise search, and database queries \cite{googleBigQueryDatabaseTool2025, lramosveaMicrosoft365Copilot2026}. In Agentic RAG, agents decide what to retrieve. Many architectures have been proposed \cite{singhAgenticRetrievalAugmentedGeneration2026,li-etal-2025-survey,subramanian2025keywordsearchneedachieving}, ranging from single-agent loops with retrieval tools to multi-agent and hierarchical designs. Deep research systems are a special form of Agentic RAG \cite{li-etal-2025-survey}. A detailed study of the architectures of agentic RAG products \cite{gleanWaldoAgenticSearch2026,awsAgenticRetrievalAugmented2025} and deep research systems \cite{GeminiDeepResearch2026, DeepResearchChatGPT2026, hadfieldHowWeBuilt2025} is beyond the scope of this paper.

\subsection{AI-Enhanced Workflows}
\label{sec:workflows}

Besides denoting an architectural MAS pattern, the term \emph{workflow} also
denotes a class of software that predates the current wave of agentic AI:
platforms for robotic process automation (RPA), workflow automation and
business process management. 
The platforms target trained end users with low-code or no-code environments in
which workflows are assembled by arranging and connecting prefabricated blocks
in a graphical user interface.  Many of these established platforms have been
extended with blocks that invoke LLM agents; these agents are
configured in dedicated agent-builder areas
\cite{uipathAgentsPromptsArguments2025,begoSuperchargeYourAgents2025}. For instance, such agents  draft 
replies to customer e-mails, summarize incoming messages or route
them to the responsible staff member.
One platform aimed at IT
professionals makes agent invocations available as modeling elements within
process models expressed in the standardized Business Process Model and
Notation (BPMN) \cite{camundaAgenticOrchestration2025}. Another is
source-available, can be self-hosted and used with arbitrary models, including
local ones, which helps to satisfy strict data protection requirements, for
example in public administration \cite{n8n-ioN8nioAIWorkflow2025}.

The workflows are typically triggered by events or schedules and then run
unattended. Systems may provide checkpoints at which humans inspect
intermediate results, either on every execution or only where a decision is
flagged as critical or falls below a confidence threshold
\cite{uipathIntelligentDocumentProcessing2025}; outside these checkpoints,
intermediate results remain uninspected.

By contrast, the LLM ecosystem offers programming frameworks for building
agentic workflows, together with visual builders layered on top of them
\cite{langchain2025,LangflowVisualLowCode2025,llamaindex2025}. These address
developers and permit more flexible MAS architectures than the prefabricated
blocks of the low-code platforms.

\subsection{Copilots}
\label{sec:copilots}

\emph{Copilot} has become an established term for LLM-integrated systems.
Our survey of commercial systems and research prototypes labeled copilots
aims to identify their common characteristics. It draws on vendor
documentation and, as this is often not very detailed, also on direct
inspection of the systems and on the findings of an
interview study with engineers building product copilots
\cite{parnin2025building}. Table~\ref{tab:copilot-evidence} summarizes the
results.

\begin{table}[t]
  \caption{Properties of Copilots according to the consulted sources}
  \label{tab:copilot-evidence}
  \footnotesize

  {\centering
  \begin{tabular}{p{4.3cm}cccc}
    \toprule
    \textbf{System} & \textbf{RW\textsuperscript{a}} & \textbf{Cont\textsuperscript{b}} & \textbf{Host\textsuperscript{c}} & \textbf{Mono\textsuperscript{d}} \\
    \midrule
    Excel Copilot, Microsoft 365 Copilot \cite{fortinMicrosoftCopilotExcel2024,spataroIntroducingMicrosoft3652023} & $\circ$ & \checkmark & \checkmark & \checkmark \\
    Dynamics 365 Copilot \cite{dynamics365copilot2023} & ? & ? & \checkmark & \checkmark \\
    SAP Joule, procurement \cite{sapJouleFunktionenBeschaffung2025} & \checkmark & \checkmark & \checkmark & \checkmark \\
    Salesforce Einstein Copilot \cite{salesforceEinsteinCopilotGA2024} & \checkmark & ? & \checkmark & \checkmark \\
    Siemens Industrial Copilots \cite{siemensSiemensIndustrialCopilot2024} & \checkmark &\checkmark & \checkmark & $+$  \\
    GitHub Copilot \cite{GitHubCopilotYour2026} & $\circ$ & \checkmark & \checkmark & $\circ$ \\
\midrule
    Product Configuration Copilot \cite{kogler2024configuration} & ? & $\checkmark$ & $\checkmark$ & $\checkmark$ \\
    ADMA Copilot \cite{panBuildingMultiAgentCopilot2024} &  $\times$   & $\times$ &  \checkmark & \checkmark \\
    Equipment Selection Copilot \cite{werheid2025designing} &\checkmark  & $\times$ & $\times$ & $+$ \\
    \bottomrule
  \end{tabular}
\par}
\vspace{2mm}
{\footnotesize
\textsuperscript{a}RW: router-worker topology.
\textsuperscript{b}Cont: user control; they confirm actions and/or can undo effects.
\textsuperscript{c}Host: tools
  are functions operating a host application.
\textsuperscript{d}Mono: monofunctional workers.
\par
  \checkmark\ stated in the source; ? not stated; $\times$ the source states otherwise;
 $\circ$ inferred from \cite{parnin2025building} or drawn from the author's own inspection;
$+$ router-worker at the top level, but workers are themselves workflow/orchestrator agents, not monofunctional.
\par}
\end{table}

\subsubsection{Commercial Copilots}

One office suite vendor describes an orchestration engine that combines the
language model with the suite's applications and the user's business data,
together with an extensible set of \emph{skills} that the assistant applies
\cite{spataroIntroducingMicrosoft3652023}. The same source states that the
user decides what to keep, modify or discard; an early demo video shows this
behavior \cite{fortinMicrosoftCopilotExcel2024}.
The same vendor's copilot for CRM and ERP is likewise a set of delimited
capabilities operating on the functions of the host applications: it drafts
replies from case history and knowledge bases, curates customer segments in
dialogue with the customer data platform, and generates product descriptions
from stored item attributes \cite{dynamics365copilot2023}.

A further CRM vendor describes the two parts of the architecture explicitly:
a reasoning engine that interprets user requests and selects from a library
of \emph{actions}, described as pre-programmed capabilities, and a set of
such actions delimited to one task each
\cite{salesforceEinsteinCopilotGA2024}. The same source describes
context-dependent single-click actions offered on the page the user is
currently on, such as summarizing an opportunity, which is the button
mechanism noted above.

In a further ERP suite the assistant contains workers that execute tasks
directly; in the procurement module these create purchase requisitions,
change purchase orders and so on, without the user opening the respective
applications \cite{sapJouleFunktionenBeschaffung2025}. The vendor's technical
documentation records each of these capabilities separately, with its own set
of application programming interfaces and its own semantic object-action
pair, the platform's established means of addressing an application function.
The worker set and the binding of each worker to its tools are thus
documented capability by capability.

An industrial vendor develops copilots along the value chain, from design
through engineering, planning, and operations to service. The architecture
distinguishes between copilots and specialized agents: users interact with
the copilot, while an orchestrator delegates tasks to specialized agents
\cite{SiemensIntroducesAI2025}. A demo shows the operations copilot being
accessed via a chat interface \cite{charbaxSiemensInsightsHub2025}. Agents
are described as understanding intent and accessing external tools and other
agents; they create plans and execute actions independently
\cite{siemensSiemensIndustrialCopilot2024}, while users retain full control
and can select which tasks to delegate to them
\cite{SiemensIntroducesAI2025}. This indicates that the copilot implements a
router-worker pattern at the level at which the user interacts with it, while
the workers---though specialized---may themselves implement the
orchestrator-worker pattern described in Section~\ref{sec:amas}.

A prominent programming copilot product installs into development
environments as a plug-in and uses their functions, e.g., for editing and
creating files or writing test routines. Product screenshots show specialized
copilot skills---the workers---which the user can invoke directly with a
slash command, thereby bypassing the copilot's intent detection
\cite{GitHubCopilotYour2026, GitHubCopilotAI2026}. In its FAQ, the vendor
states the division of responsibility as a design principle, noting that the
product is called \emph{copilot} rather than \emph{autopilot} and that
generated code is not intended to be compiled or executed before the
developer has reviewed it \cite{GitHubCopilotYour2026}.

\subsubsection{Research Prototype Copilots}

 A product configuration copilot
provides a chat interface for a product configurator. It
translates natural-language user input into a format the constraint resolver
can process automatically, and phrases the results in natural language for
the user \cite{kogler2024configuration}. This copilot operates an existing
application, and the user can see and correct the effect of the operation
immediately. With only one supported intent, the setup is too simple to
require routing.
An agricultural copilot operates  an  Agricultural Data Management and Analytics (ADMA) platform. Its controller agent 
 infers the user's intent, determines which of a
platform's tools and processing steps the intended analysis requires,
parametrizes them, and coordinates their execution following a predefined meta-program graph
\cite{panBuildingMultiAgentCopilot2024}. This resembles the
orchestrator-worker architecture of Subsection~\ref{sec:amas} more than the
router-worker one.

A copilot for manufacturing equipment selection follows the router-worker
pattern at the level at which the user interacts with it, while one of its
workers is internally a workflow of successive model invocations: the
suitability of a selection is checked by a further model invocation rather
than by the user, who is consulted only when that check fails
\cite{werheid2025designing}. This copilot accesses knowledge, but does not
operate further application software.

\subsubsection{Characteristics of Copilots}

Typical copilots operate existing application software, visibly performing
steps previously done by hand; commercial ones are frequently integrated
directly into the host product. Architecturally, they follow a router-worker
pattern: the user states a goal in a chat window, the copilot detects the
intent, and specialized workers carry it out. Because the effect of each
operation is visible, the user can keep it or refine it in further turns.
Some copilots additionally propose contextually appropriate actions, for
example as buttons triggering the corresponding intent.

\subsubsection{Shift of Labels}
The survey indicates that vendors use the terminology more consistently than
researchers do. More recently, vendors have begun to brand their AI products
as agents, in addition to or instead of as copilots, and this shift of labels
coincides with a change in architecture. One ERP provider explicitly
announces its AI agents as a move from ``just AI copilot'' to ``AI
orchestrator'' \cite{bownJouleAgentsHow2025}; a comparable statement is made
with respect to industrial copilots \cite{SiemensIntroducesAI2025},
and a further vendor mentions ``dynamic multi-step plans'' within an agent
platform \cite{salesforceeinsteincopilot}. The programming copilot has been
augmented by an ``agentic mode'' \cite{CodingHarnessGitHub2026}.

\begin{table*}[ht]
  \caption{Forms of LLM-integrated applications}
  \label{tab:systems}
  \footnotesize
\setlength{\tabcolsep}{3pt}
  {\centering
  \begin{tabular}{p{3.3cm}p{4.0cm}p{2.8cm}p{2.7cm}p{2.1cm}p{2.0cm}}
    \toprule
    \textbf{Form or \textit{System}} & \textbf{Architecture\textsuperscript{a}} & \textbf{User control} & \textbf{Agent calls per task\textsuperscript{b}} & \textbf{Tool use} & \textbf{Sources} \\
    \midrule

   LLM chats\textsuperscript{c} &
    multifunctional single agent &
    each answer &
    one  &
    few &
    \cite{schwartzChatGPTJustAccidentally2024,openaiIntroducingGPT52025} \\

    Custom agents &
    mono\-functional single agent &
    each answer &
    one  &
    at most one &
  \cite{quorainc.PoeFastHelpful2025,zhaoGPTsWindowShopping2024,openaiacademyBuildingCustomGPTs2025} \\

    Agentic RAG  &
   ? &
    each answer &
    ?   &
    retrieval tools &
 \cite{googleBigQueryDatabaseTool2025,subramanian2025keywordsearchneedachieving,lramosveaMicrosoft365Copilot2026} \\

    RAG, narrow  &
    mono\-functional single agent &
    each answer &
    one  &
    none &
   \cite{lewis2020retrieval} \\

    AI-enhanced workflows &
    workflow &
    optional checkpoints  &
    fixed (process) &
    fixed (process) &
  \cite{uipathAgentsPromptsArguments2025,begoSuperchargeYourAgents2025,camundaAgenticOrchestration2025,n8n-ioN8nioAIWorkflow2025, uipathIntelligentDocumentProcessing2025} \\

    Copilots &
    router-worker &
    each action result &
    two &
    host functions &
     \cite{spataroIntroducingMicrosoft3652023,dynamics365copilot2023,sapJouleFunktionenBeschaffung2025,salesforceEinsteinCopilotGA2024,siemensSiemensIndustrialCopilot2024,GitHubCopilotYour2026} \\

    Coding agents &
    ReAct  loop &
    final result only &
     open, unbounded &
    extensive &
\cite{oroszHowCodexBuilt2026,HowClaudeCode2026,WhatAgenticCoding2026,phdComponentsCodingAgent2026,CodingHarnessGitHub2026} \\
\midrule

    \textit{Equipment selection copilot}  & router, workflow-type worker  & final result only  & bounded & no tools  & \cite{werheid2025designing} \\

    \textit{Agricultural copilot} &
    orchestrator-worker &
    final result only &
     open, bounded &
    platform functions &
\cite{panBuildingMultiAgentCopilot2024} \\

    \textit{Research Agent} &
    planner-executor, auton. subagents &
    final result only &
     open, unbounded &
    retrieval tools  &
\cite{hadfieldHowWeBuilt2025} \\

    \bottomrule
  \end{tabular}
  \par}
  \vspace{2mm}
  {\footnotesize
  \textsuperscript{a}Architecture: pattern of Section~\ref{sec:amas}.  \textsuperscript{b}Task:  one unit of work the system is given to complete: a user request,
  a conversational turn, or a triggered run of a process.
  \textsuperscript{c}LLM chat: excluding the deep research function.
  \par}
\end{table*}

\subsection{Coding Agents}
Among the emerging forms of AMAS applications that share a common architecture are coding agents for software development.
They are built around an agentic loop that delegates work to subagents \cite{oroszHowCodexBuilt2026, HowClaudeCode2026, WhatAgenticCoding2026, phdComponentsCodingAgent2026,anthropicBuildingEffectiveAgents2024}. A coding copilot (GitHub Copilot) has since been extended with an agent mode that follows the same pattern \cite{CodingHarnessGitHub2026}.

\subsection{Planner-Executor Example System}

A further AMAS example is Anthropic's Research feature. Its developers describe it as a multi-agent architecture in which a lead agent analyzes the query, devises a research strategy and delegates subtasks to subagents that run in parallel, then synthesizes their findings and decides whether further research is needed; if so, it spawns additional subagents or revises its strategy \cite{hadfieldHowWeBuilt2025}.
Although its developers name it an orchestrator-worker AMAS, this comes close to the planner-executor pattern as defined in this paper.  
 The correspondence is not exact: the subagents are themselves tool-using agents that search, evaluate intermediate results and reformulate their own queries, so they are considerably more autonomous than executors in the classical sense; and planning is interleaved with execution rather than completed up front.

\subsection{Summary of Selected Systems}
\label{sec:systems}

To illustrate the descriptions given above, Table~\ref{tab:systems} summarizes
the forms, together with the system instances for which this study found no
established form. For each entry it records four properties: the architectural
pattern, the point at which a human user intervenes, the number of agent calls
per task, and the tools available to the agents. The entries combine
commercial products, for which vendor documentation is the principal source,
with research prototypes, for which peer-reviewed descriptions are available; see Section~\ref{sec:limitations} for discussion.

\section{Discussion}
\label{sec:discussion}

This study asks which structurally similar forms of LLM-integrated systems have already emerged, given that interest in and development of these systems began less than four years before this study. It investigates whether the labels applied to LLM-integrated systems denote actual forms beyond mere branding, and, where they do, what characteristics constitute each label.
The study focuses on structural dimensions to characterize the systems, including system architecture, degree of user control, number of agent calls per task, and tool use, as such characteristics make it possible to describe newly developed systems in new application areas.

\subsection{Scope and Granularity}
\label{sec:granularity}
The study does not aim to provide a taxonomy that sorts systems into separate,
mutually exclusive categories, nor to map a fine-grained design space for
developers or to serve as an analytical instrument for detailed architectural
research. It characterizes LLM-integrated systems at a general level, which is
what its purpose requires: to provide an accessible framework for
practitioners, students, and newcomers to the field, and to help position and
label systems using established, common terms.

\subsection{Labels and Architectural Distinctions}
The survey indicates that vendors use the label copilot consistently, as detailed in Section~\ref{sec:copilots}. The subsequent shift to \emph{agent} is
accordingly more than a change in fashion, as it marks a move from
AI-selected single steps that the user confirms to AI-planned multi-step
execution of which the user sees only the outcome.
The survey also found that the coding agents of four major providers are similar along the structural dimensions examined. For RAG in the narrow sense, the characteristics were derived from the original formulation \cite{lewis2020retrieval}; the survey did not investigate to what extent products or prototypes branded as RAG exist and conform to this definition. 
The study introduces its own label, AI-enhanced workflow, to the author's knowledge there is not yet a common label for this form.

\subsection{Nested and Hybrid Architectures}

LLM-integrated systems are complex and vary in their details; MAS in particular can be designed very flexibly. Diverse, hybrid, and nested architectures are possible and common.

The Equipment Selection Copilot \cite{werheid2025designing} is an example: at the level the user interacts with, it is a router-worker copilot, but one of its workers is internally a workflow of several chained model invocations.

As agentic systems become more common, more such nested architectures are expected to appear, as copilot-type and chat-type systems become enhanced with MAS-type workers and tools. The chat or copilot pattern acts at the user-interface level, which starts the agentic workers or tools autonomously, in cooperation with, or with confirmation from, the user.
For these cases, it is suggested that the classification be drawn at the user-facing level, with the internal nesting recorded as a property of the respective workers, rather than treated as a separate system form.

\subsection{Limitations and Future Work}
\label{sec:limitations}

Systems were characterized by a single author from published descriptions;
for several commercial systems, these, necessarily, are vendor sources that disclose only a certain level of technical detail; in some cases, the characterization is based only on what the vendor states. The corpus is
illustrative rather than representative, and no frequency claims follow
from it. Independent second coding, and characterization
against technical rather than promotional sources where these become
available, would strengthen the observations.

The descriptions were developed over a period in which the field changed substantially: products marketed as copilots acquired capabilities now marketed as agents, and AMAS moved from research prototypes to commercial offerings. New forms of LLM-integrated systems keep emerging, for example locally hosted, user-configurable personal assistants that read and write files, run commands, and browse the web on the user's own machine \cite{ClaudeCoworkGets2026}. Extending the study to such new forms, and to further existing forms such as Deep Research and Agentic RAG, is a natural direction for future work.

Matching forms of LLM-integrated systems to other kinds of AI-based system not built on LLMs could help clarify their relation to the more recent forms of agentic AI; transactional chatbots \cite{baezChatbotIntegrationFew2020} and the expert systems of the 1990s \cite{hayes-rothBuildingExpertSystems1983} are two examples.

\section{Conclusion}

Many terms for LLM-integrated systems---chat, custom agent,
agentic RAG, RAG, copilot, coding agent, etc.---are
applied routinely in practice but have, with exception of RAG, not been
described in the research literature in a way that states uniformly how
the systems behind them are built.

This paper has described six forms of such systems in full, and a seventh in part,
illustrated by a survey of 22 systems characterized from vendor sources,
own inspection, and video material, and informed by research
publications, survey papers, and other secondary sources. Each form is
characterized by its architecture, by the role left to the human user,
by the number of agent calls per task, and by the tools its agents use.

The classification is offered as a means of relating these systems to
one another: to make the field more accessible for newcomers and in
teaching, to give practitioners a basis for comparing technologies, and
-- as a theoretical and conceptual contribution---to introduce the
emerging terminology into the research literature so that studies can
refer to it.

\end{document}